\documentclass[10pt,twocolumn,letterpaper]{article}

\usepackage{cvpr}
\usepackage{times}
\usepackage{epsfig}
\usepackage{graphicx}
\usepackage{amsmath}
\usepackage{amssymb}

\usepackage{times}
\usepackage{booktabs}
\usepackage{amsthm}
\usepackage{balance}
\usepackage{algorithmic}
\usepackage{mathtools}
\usepackage{enumerate}
\usepackage{balance}
\usepackage{mathptmx}

\usepackage{algorithm}

\usepackage{booktabs}
\usepackage{caption}
\usepackage{etoolbox}
\usepackage{fancyhdr}
\usepackage{float}
\usepackage{graphicx}
\usepackage{paralist}
\usepackage{rotating}
\usepackage[detect-weight]{siunitx}
\usepackage{tabularx}
\usepackage{url}
\usepackage{xspace}
\usepackage{bm}
\usepackage[bottom]{footmisc}%
\usepackage{algorithm}

\usepackage{amssymb, amsmath} 

\usepackage[pagebackref=true,breaklinks=true,letterpaper=true,colorlinks,bookmarks=false]{hyperref}

 \cvprfinalcopy 

\graphicspath{{./}{./figures/}}
\def\Vect#1{{\boldsymbol{#1}}}
\def\Mat#1{{\boldsymbol{#1}}}
\renewcommand{\Vect}[1]{\boldsymbol{\mathbf{#1}}}
\renewcommand{\Vec}[1]{\boldsymbol{\mathbf{#1}}}
\renewcommand{\Mat}[1]{\boldsymbol{\mathbf{#1}}}

\floatstyle{ruled}
\newfloat{algorithm}{tb!}{loa}[chapter]
\floatname{algorithm}{Algorithm}
\makeatletter
\renewcommand\fs@ruled{\def\@fs@cfont{\bfseries}\let\@fs@capt\floatc@ruled
  \def\@fs@pre{\hrule height1pt \kern10pt}%
  \def\@fs@post{\kern8pt \hrule height1pt \relax}%
  \def\@fs@mid{\kern8pt \hrule height1pt \kern10pt}%
  \let\@fs@iftopcapt\iftrue}
\makeatother

\newcommand{\tr}{\operatorname{Tr}}

\ifcvprfinal\fi
\begin{document}

\title{Optimized Kernel-based Projection Space of Riemannian Manifolds}

\author{Azadeh Alavi\\
University of Maryland\\
\and
Vishal M Patel\\
Rutgers University\\
\and
Rama Chellappa\\
University of Maryland\\
}

\maketitle

\begin{abstract}
\begin{quote}
Recent advances in computer vision suggest that encoding images through Symmetric Positive Definite (SPD) matrices can lead to increased classification performance. Taking into account manifold geometry is typically done via embedding the manifolds in tangent spaces, or Reproducing Kernel Hilbert Spaces (RKHS). Recently it was shown that projecting such manifolds into a kernel-based random projection space (RPS) leads to higher classification performance. 
In this paper, we propose to learn an optimized projection, based on building local and global sparse similarity graphs that encode the association of data points to the underlying subspace of each point.
To this end, we project SPD matrices into an optimized distance preserving projection space (DPS), which can be followed by any Euclidean-based classification algorithm. Further, we adopt the concept of dictionary learning and sparse coding, and discriminative analysis, for the learned DPS on SPD manifolds. 
Experiments on face recognition, texture classification, person re-identification, and virus classification demonstrate that the proposed methods outperform state-of-the-art methods on such manifolds.
\end{quote}
\end{abstract}

\vspace{-1ex}
\section{Introduction}
\label{sec:introduction}
In recent years, covariance matrices have been successfully used as image and video descriptors \cite{alavi2014random, alavi2013relational, phillips2000feret,veeraraghavan2005matching},
as they are known to provide compact and informative feature descriptions \cite{DPM_CVPR_2011,cherian2012efficient}.
Non-singular covariance matrices are Symmetric Positive Definite (SPD), and form connected Riemannian manifolds \cite{lui2011tangent}.
As such, to learn appropriate classifiers, the Riemannian geometry needs to be considered \cite{veeraraghavan2005matching}. 
Affine Invariant Riemannian Metric (AIRM) can be considered as one of the most widely used similarity measures for SPD matrices \cite{phillips2000feret}. 
The AIRM induces the Riemannian structure which is invariant to inversion and similarity transforms.
Despite its properties, methods using this approach have to deal with computational challenges, as non-linear operators are involved.
To address the issue discussed above, two lines of research have been proposed in the literature:
(1)~embedding manifolds into manifold tangent spaces \cite{Pennec_jmiv06,Brodatz_Dataset,veeraraghavan2005matching,elhamifar2013sparse}; and
(2) embedding via the reproducing kernel Hilbert space (RKHS) \cite{sivalingam2010tensor, harandi2012sparse, alavi2013relational}.
Approaches under the first category in effect map points on manifolds to Euclidean spaces,
thereby enabling the use of existing Euclidean-based learning algorithms,
but at the cost of disregarding some of the manifold structure.
Approaches under the second category address this by first mapping points on the manifold into RKHS, which can be considered as a high dimensional Euclidean space.
Training data can be used to define a space that preserves manifold geometry \cite{harandi2012sparse}.
The downside is that existing Euclidean-based learning algorithms need to be kernelised, which may not be trivial.
Furthermore, the resulting methods can still have high computational load, making them impractical in more complex scenarios.

To address the discussed above drawbacks, the random projection method over RKHS (ROSE) \cite{alavi2013relational}, which combines the main advantage of tangent space approaches with the high accuracy provided by kernel space methods was recently proposed. By adapting a recent idea from techniques specifically designed for learning tasks in very large image datasets \cite{guo2010action,lang1999fundamentals}, image representations are mapped into a reduced space wherein the similarities are still well-preserved \cite{lang1999fundamentals}.
The additional advantage of using such a method is that, the computational burden is considerably decreased while maintaining good classification performance. In other words, the proposed approach employs a mapping technique to create a space that preserves manifold geometry well, and can be considered as Euclidean.

However, the methods outlined above have some disadvantages. First, the dimensionality of the mapping matrix is unknown, and one can only rely on the historical data to compute appropriate dimensionality for each given application. In scenarios where the number of data is limited, deciding on proper dimensionality might not be feasible.
In addition, as the name indicates, ``Random Projection" is based on random sampling of the data. Hence, in cases where the data size is sufficiently large, finding optimum random projection would require a very large number of random selections. As such, finidng an optimum random projection can become impractical.  \par
\textbf{Contributions.} In this paper, we propose to  

\begin{itemize} 
\item
 Learn the structure of the projection map by taking advantage of the concept of self-expressiveness property from subspace-sparse representation theory \cite{elhamifar2013sparse}. Thus, we construct both local and global sparse similarity graphs, whose nodes are
connected to each other iff they correspond to points from the same subspace. 
To this end, we embed the SPD matrices into RKHS via the Stein Divergence Kernel 
\footnote{We note, there are other SPD kernels that could be employed, such as Gaussian kernel \cite{arsigny2006log, jayasumana2015kernel}, or Stein based kernel on infinite-dimensional Covariance Descriptors \cite{harandi2014bregman}. However to keep the results comparable with previous works we use Stein Divergence Kernel in this paper.}, \cite{sra2011generalized,alavi2014random}. 
Then we invoke the subspace-sparse representation theorem, and combine it with the random projection method over RKHS (ROSE). As such, we learn an appropriate projection which maps the embedded points into DPS.
\item
This is followed by utilizing any existing Euclidean-based learning algorithms, such as SVM. This does not violate the underlying structure of the SPD manifolds, as the proposed DPS preserves the manifold geometry. In addition, this mapping mitigates the drawbacks of the random projection on SPD manifolds method.
Experiments on several vision tasks (person re-identification, face recognition, texture recognition, and virus classification),
show that the proposed approach outperforms the random projection method over RKHS (ROSE) and several other state-of-the-art methods.
\item
Next, we adopt discriminative analysis (DA), and dictionary learning (DL) on the learned DPS of SPD manifolds. We further enhance the discriminative power of the method and extend the comparison of DPS-based with RKHS-based methods on SPD manifolds. Finally, we compare the performance of the proposed methods with other DA, and DL methods \cite{harandi2012sparse, shirazi2013graph,harandiriemannian} on such manifolds.
\end{itemize}

Rest of the paper is organized as follows:
Section~\ref{sec:LogDetDivergance} provides a brief overview of the manifold structure and its associated kernel function. Details of the Random Projection on RKHS of SPD manifolds are provided in Section~\ref{sec:KLSH}.
We then present the proposed approach in Section~\ref{sec:DPS}. Experimental results are provided in Section~\ref{sec:EXP}, and Section~\ref{sec:Conc} concludes the paper with a brief summary and discussion. 

\section{Manifold structure and Stein Divergence Kernel}
\label{sec:LogDetDivergance}

In this section, we first brief the SPD manifold geometry, and then introduce one of the SPD manifold kernels known as Stein Divergence \cite{sra2011generalized}.

Consider $\mathbb{X}  = \{ {\Mat{X}}_1 \dots {\Mat{X}}_n\}\in \mathcal{S}_{d}^{+}$ as a set of non-singular $d$ by $d$ covariance matrices, which are naturally SPD matrices. These matrices belong to smooth differentiable topological space, known as SPD manifolds.  
In this work, we endow the SPD manifold with the AIRM over its tangent spaces to induce the Riemmanian structure into it \cite{Turaga_2010_Book_Chapter}.
We define the logarithm operation which maps points over the manifold to a tangent space as:
\begin{equation}
    \operatorname{log}_{\Mat{X_i}} {\Mat{X_j}} = {{\Mat{X_i}}^{\frac{1}{2}} \operatorname{logm}({ {\Mat{X_i}}^{-\frac{1}{2}}  \Mat{X_j}  {\Mat{X_i}}^{-\frac{1}{2}}    }) \Mat{X_i}^{\frac{1}{2}}},
    \label{eqn:AIRM_log}
\end{equation}
where $\Mat{X_i}, \Mat{X_j} \in \mathcal{S}_d^{+}$, $\Mat{X_i}$ is the point where the tangent space is located and $\Mat{X_j}$ is the point that we would like to map into the tangent space $\mathcal{T}_{\Mat{X_i}}\mathcal{M}$; 
$\operatorname{logm} ( \cdot )$ is the matrix logarithm.
The inverse function of this maps from points over a particular tangent space into the manifold defined via
\begin{equation}
    \operatorname{exp}_{\Mat{X_i}} {\Mat{y}} = {{\Mat{X_i}}^{\frac{1}{2}} \operatorname{expm}({ {\Mat{X_i}}^{-\frac{1}{2}} \Mat{y}  {\Mat{X_i}}^{-\frac{1}{2}}    }) \Mat{X}^{\frac{1}{2}}},
    \label{eqn:AIRM_exp}
\end{equation}
where $\Mat{X_i} \in \mathcal{S}_d^+$ is the tangent pole; 
$\Mat{y} \in \mathcal{T}_{\Mat{X_i}}\mathcal{M}$ is a point in the tangent space $\mathcal{T}_{\Mat{X_i}}\mathcal{M}$;
$\operatorname{expm} ( \cdot )$ is the matrix exponential. Using these definitions, we now define the shortest distance between two points over the manifold.
The geodesic distance \cite{phillips2000feret}, which is defined as the minimum length of the curvature path that connects between two points, can be written as:
\begin{equation*}
     \operatorname{d}_g^2{(\Mat{X_i},\Mat{X_j})} = \operatorname{trace}({{\operatorname{logm}^2} ({  \Mat{X_i}^{-\frac{1}{2}} \Mat{X_j}  {\Mat{X_i}}^{-\frac{1}{2}}  })     }).
    \label{eqn:AIRM_dg}
\end{equation*}

\subsection{Stein Divergence Kernel}
\label{subsec:LogDetDiverganceKernel}
In this work, we employ the recently introduced Stein divergence \cite{sra2011generalized} to calculate the distance between points on Riemannian manifolds. Stein divergence has an advantage of computational speed compare to geodesic distance, and is defined as follows
\begin{equation}
    J_{\phi }(\Mat{X},\Mat{Y}) \triangleq  \log \left( \det \left( \frac{\Mat{X}+\Mat{Y}}{2}\right) \right)
    - \frac{1}{2}  \log \left( \det \left( \Mat{X}\Mat{Y} \right) \right) .
    \label{eqn:Stein_Div}
\end{equation}%
Stein Divergence kernel is then defined based on the symmetric Stein divergence dissimilarity function
    \begin{equation}
        \operatorname{K}(\Mat{X},\Mat{Y}) = e^{-\sigma J_{\phi }(\Mat{X}\hspace{-0.25ex},\Mat{Y})} ,
        \label{eqn:SDKernel}
    \end{equation}
    where  \mbox{\small $\Mat{X},\Mat{Y} \in \mathcal{S}_{++}^d$} \cite{sra2011generalized}.  Consider $\{\Mat{X}_1 \dots \Mat{X}_n\}$ to be a set of matrices on the SPD manifold, the $n \times n$ matrix $\mathbb{K}_\sigma (i,j) = \operatorname{K}(\Mat{X}_i,\Mat{X}_j)$ defined in (\ref{eqn:SDKernel}) is positive definite if the following condition is satisfied \cite{Sra_JMLR_2012}
    \begin{equation}
		\sigma \in \left\{\frac{1}{2},\frac{2}{2},...,\frac{n-1}{2}\right\}.
		\label{eqn:SDKernelCond}
		\end{equation}
Therefore, if (\ref{eqn:SDKernelCond}) is satisfied, (\ref{eqn:SDKernel}) forms a Riemannian Kernel. 


\section{Random Projection over RKHS}
\label{sec:KLSH}

The recently proposed approach known as ROSE, addresses image classification tasks originally formulated on the manifold by embedding them into a random projection space (Euclidean Space), while still respecting the underlying manifold structure.  Random projection is an approximation method for estimating he dis-similarity between pairs of point in a high-dimensional space \cite{achlioptas2003database}.
Basically, the projection of a point $\Vec{u} \in \mathbb{R}^d$ can be done via a set of randomly generated hyperplanes $\{ \Vec{r}_1 \dots \Vec{r}_k \} \in \mathbb{R}^d$
\begin{equation}
	{f}(\Vec{u}) = \mathbf{u}^{\top} \mathbf{W},
	\label{eqn:KLSH_mapping}
\end{equation}
where $ \mathbf{W} \in \mathbb{R}^{d \times k}$ is the matrix wherein each column contains a single hyperplane $\Vec{r}_i$;
${f}( \cdot )$ is the mapping function which maps any point in $\mathbb{R}^d$ into a reduced space $\mathbb{R}^k$.
According to the Johnson-Lindenstrauss lemma \cite{achlioptas2003database}, this mapping approximately preserves the pairwise distance between two points in the projected space.

Despite the popularity of the Johnson-Lindenstrauss lemma, many proposed methods restrict the distance function to $l_p$ norm, the Mahalanobis metric, or the inner product \cite{charikar2002similarity,datar2004locality,kulis2012kernelized}, which makes them non adoptable for non-Euclidean spaces. 
Recently, Kulis and Grauman \cite{kulis2012kernelized} proposed a method that allows the distance function to be evaluated over RKHS.
Thus, it is possible to apply the lemma for any arbitrary kernel $\mathbb{K}(i,j)={K}(\Vec{X}_i,\Vec{X}_j)={\phi}(\Vec{X}_i)^{T}{\phi}(\Vec{X}_j)$ for an unknown embedding ${\phi}( \cdot )$ which maps the points to a Hilbert space $\mathcal{H}$ \cite{kulis2012kernelized}. 
This approach makes it possible for one to construct a random projection space over an SPD manifold, where the manifold structure is well-preserved.

As such, ROSE proposed  to first map all the points over the manifold into RKHS via the Stein Divergence SPD kernel function ${\phi}( \cdot )$, and then map all the points in the RKHS ${\phi}(\Vec{\Mat{X_i}}) \in \mathcal{H}$  into a random projection space $\mathbb{R}^k$ with the aid of the recently introduced Kernelized Sensitive Hashing approach \cite{kulis2012kernelized}.
To achieve this, they follow Kulis-Grauman method \cite{kulis2012kernelized} by randomly generating a set of hyperplanes over the RKHS $\{\Vec{r}_1 \dots \Vec{r}_k\} \in \mathcal{H}$ which are normalized to be approximately Gaussian.
As the embedding function ${\phi}( \cdot )$ is considered as unknown, then the generation process is done indirectly via a weighted sum of the subset of the given training sets.

To this end, consider each data point ${\phi}(\Mat{X}_i)$ from the training set as a vector from some underlying distribution $D$ with unknown mean $\boldsymbol{\mu}$ and unknown covariance $ \Sigma$. Let $S$ be a set of $t$ IID training exemplars from $D$, then $\Vec{z}_t = \frac{1}{t}\sum_{i\in S} {\phi}(\Mat{X}_i)$ is defined over $S$. 
According to the central limit
theorem for sufficiently large $t$, the random vector $\tilde{\Vec{z}_t} = \sqrt t (\Vec{z}_t - \boldsymbol{\mu})$ is distributed as a multi-variate Gaussian $\mathcal{N}(\boldsymbol{\mu},  \Sigma)$ \cite{Schwartz_ETHZ_short}.
Then if the whitening transform: $\Vec{r}_i = {\Sigma} ^{-\frac{1}{2}} \tilde{\Vec{z}}_t$ is applied which yields $\mathcal{N}(0,{\Sigma})$ distribution in the Hilbert space $\mathcal{H}$. Therefore the $i$-th value of each vector in the random projection space is defined as
\begin{equation}
	{\phi}(\Vec{X}_i)^{T} \Sigma ^{-\frac{1}{2}} \tilde{\Vec{z}}_t.
	\label{eqn:KLSH_embedding}
\end{equation}
As mean $\boldsymbol{\mu}$ and covariance ${\Sigma}$ are unknown, they need to be estimated from the sample data. 
So a set of $p$ objects from the dataset is chosen to form the first $p$ items of the database: ${\phi}(\Vec{X}_1) \dots {\phi}(\Vec{X}_p)$ . Then the mean is estimated as $\boldsymbol{\mu} = \frac{1}{p}\sum_{i=1}^{p} {\phi}(\Vec{X}_i)$, and the covariance matrix $ \Sigma$ is also estimated using the $p$ centred samples.
Equation (\ref{eqn:KLSH_embedding}) can be computed using the Kernel PCA \cite{kulis2012kernelized}.

To summarize, the columns of the mapping matrix $\Mat{W}$ from (\ref{eqn:KLSH_mapping}), are calculated using
\begin{equation}
	{\Vec{w}}_i = {\Mat{K}}^{\frac{1}{2}}\left ( \frac{1}{t}\mathbf{e}_s-\frac{1}{p}\mathbf{1}\right ),
	\label{eqn:KLSH_K_Half}
\end{equation}
where $\Mat{K}$ is the kernel matrix of Training data, $\mathbf{1}$ is a vector of all ones, and $\mathbf{e}_s$ is a sparce vector with ones in the entries corresponding to the randomly selected positions \cite{kulis2012kernelized}.  
\section{Distance Preserving Projection Space for SPD Manifolds}
\label{sec:DPS}
In this section, we first introduce the proposed algorithm for learning the optimum Distance Preserving Projection on SPD manifolds (DPS). Then, we detail the steps for performing Graph Embedding Discriminative Analysis and Dictionary Learning over the proposed DPS of SPD manifolds.
\par
\noindent
The key idea for learning the optimized DPS is inspired by the 'self-expressiveness property of the data' from subspace sparse clustering method \cite{elhamifar2013sparse}. To be specific, we propose to model ${\Mat{K}}^{\frac{1}{2}} $ structure through appropriately learned symmetric sparse similarity graphs. First, let us re-write (\ref{eqn:KLSH_K_Half}) as follows:
\begin{equation}
{\Mat{W}} = \alpha {\Mat{K}}^{\frac{1}{2}} \Mat{E}
\label{eqn:self}
\end{equation} 
, where $\alpha = \frac{\left (p-t\right )}{t}$, and $ \Mat{E} \in \mathbb{R}^{n \times t}$ is a $ \left \{ 0,1 \right \}$-matrix.
It is widely acknowledged that 
even in cases where the number of collected samples is significantly smaller than the dimensionality of the data, the data set that needs to be processed usually exhibit significant structure, which often can be captured using sparsity models \cite{bahmani2013greedy}. 
More specifically, it is shown that among the large sum possible representations of a given data point, in terms of other points, a sparse representation corresponds to selecting a few points from the same subspace \cite{elhamifar2013sparse}. 
\par
\noindent
As such, we learn the representing sparse codes for $\Mat{K}^{\frac{1}{2}}$ by formulating a local self-expressive dictionary as follows:
\begin{equation}
min{\left(\left \| \Vec{l}_{i} \right \|_1 \right)} ~s.t~ {\bar{\Vec{k}}}_i = \Mat{K^{\frac{1}{2}}}\Vec{l}_{i},~\Mat{L}_{ii} = 0.
\label{eqn:L}
\end{equation}
, where ${\Mat{K}}^{\frac{1}{2}} = \left [{\bar{\Vec{k}}}_1~{\bar{\Vec{k}}}_2~\cdots ~{\bar{\Vec{k}}}_n \right ] $, and $\Mat{L} = \{ \Vec{l}_1, \Vec{l}_2,~\cdots ~, \Vec{l}_n\}$ represents sparse codes.
Although the sparse solution to (\ref{eqn:L}) captures valuable information about the relation between points, it is known to have limitations when it comes to dealing with points near the intersection of two subspaces \cite{elhamifar2013sparse}. To further enhance the method and resolve these limitations, we also encode the similarities between data points using the sparse solution of the global self-expressive dictionary, as follows: 
\begin{equation}
min{\left(\left \| \Mat{G} \right \|_1 \right)} ~s.t~ \Mat{K^{\frac{1}{2}}} = \Mat{K^{\frac{1}{2}}}\Mat{G},~Diag(\Mat{G}_{ii}) = 0.
\label{eqn:G}
\end{equation}
\noindent
, where the sparse solutions to (\ref{eqn:L}) and (\ref{eqn:G}) can be efficiently computed by employing convex programming tools \cite{boyd2004convex}.
\par
While the sparse solutions $\Mat{L} = \begin{bmatrix}
\Vec{l}_1 & \Vec{l}_2 & \cdots & \Vec{l}_n 
\end{bmatrix} $ and $\Mat{G} = \begin{bmatrix}
\Vec{g}_1 & \Vec{g}_2 & \cdots & \Vec{g}_n 
\end{bmatrix} $, ideally correspond to subspace sparse representations of the data points, they are not necessary symmetric. However, it is clear that if ${\bar{\Vec{k}}}_i$ and ${\bar{\Vec{k}}}_j$ are members of the same subspace, i.e. $ \left | \Mat{L}_{ij} \right | >0$ (or $ \left | \Mat{G}_{ij} \right | >0$), then ${\bar{\Vec{k}}}_j$ and ${\bar{\Vec{k}}}_i$ must also be members of the same subspace, i.e respectively $ \left | \Mat{L}_{ji} \right | >0$ (or $ \left | \Mat{G}_{ji} \right | >0$). As such, we model local and global sparse symmetric similarity graphs $L = \left(\Mat{V}^l,\Mat{E}^l,\Mat{W}^l \right)$ and $G = \left(\Mat{V}^g,\Mat{E}^g,\Mat{W}^g \right)$, whose nodes resemble data points; where there is no edge between nodes that corresponds to points in different subspaces.
\par
\noindent
As such, we define the symmetric sparse similarity graphs as follows:
\begin{equation}
\Mat{W^{l}}_{(i,j)} =
\left\{
\begin{matrix}
1, & \mbox{if} \; \Mat{{E}^l}_{ij} > 0~or~ \Mat{{E}^l}_{ji} > 0\\
0, & \mbox{otherwise}
\end{matrix}
\right.
\label{eqn:Wl}
\end{equation}%
\begin{equation}
\Mat{W^{g}}_{(i,j)} =
\left\{
\begin{matrix}
1, & \mbox{if} \; \Mat{{E}^g}_{ij} > 0~or~ \Mat{{E}^g}_{ji} > 0\\
0, & \mbox{otherwise}
\end{matrix}
\right.
\label{eqn:Wg}
\end{equation}%
\par
\noindent
Considering (\ref{eqn:L}) and (\ref{eqn:G}), we reformulate (\ref{eqn:Wl}) and (\ref{eqn:Wg}) as follows:
\begin{equation}
\Mat{W^{l}}_{(i,j)} =
\left\{
\begin{matrix}
1, & \mbox{if} \; \left | \Mat{L}_{ij}\right | > 0 ~or~ \left | \Mat{L}_{ji}\right | > 0\\
0, & \mbox{otherwise}
\end{matrix}
\right.
\label{eqn:Wl2}
\end{equation}%
\begin{equation}
\Mat{W^{g}}_{(i,j)} =
\left\{
\begin{matrix}
1, & \mbox{if} \; \left | \Mat{G}_{ij}\right | > 0 ~or~ \left | \Mat{G}_{ji}\right | > 0\\
0, & \mbox{otherwise}
\end{matrix}
\right.
\label{eqn:Wg2}
\end{equation}%
\par
\noindent
Finally, Finally, to lean optimum distance preserving projection, we merge \ref{eqn:self}, \ref{eqn:Wl2}, and \ref{eqn:Wg2} \footnote{We can also form a local-based projection by merging (\ref{eqn:self}) and (\ref{eqn:Wl2}):
\begin{equation}
{\Mat{W^l}} = {\Mat{K}}^{\frac{1}{2}} \left ( \alpha \Mat{W^l} \right) 
\label{eqn:selfl}
\end{equation}
\noindent
, or similarly, formulate a global-based projection by merging (\ref{eqn:self}) and (\ref{eqn:Wg2}).
In experiment section we show that including both local and global information leads to improved classification performance.}:
\begin{equation}
{\Mat{W}} = {\Mat{K}}^{\frac{1}{2}} ( \begin{bmatrix} \left ( \alpha \Mat{W^l} \right) & \left( \beta \Mat{W^g} \right) 
\end{bmatrix} )
\label{eqn:self2}
\end{equation}
\par
To this end, we project each SPD matrix $\Mat{X}_i$ into distance preserving space (H-DPS) through bellow formulation:
\begin{equation}
\Hat{\Vec{k}_i} = \Vec{k}\left(\Mat{X}_i,\mathbb{X} \right){\Mat{K}}^{\frac{1}{2}} \begin{bmatrix} \left ( \alpha \Mat{W^l} \right) & \left( \beta \Mat{W^g} \right)
\end{bmatrix}
\label{eqn:DPSMap}
\end{equation}
\noindent
We refer interested readers to \cite{alavi2014random,kulis2012kernelized} for more detail.
\par
\noindent
In experiment section, we show that classification on the learned projection space outperforms other state of the art methods.

\subsection{Graph Embedding Discriminative Analysis}
In this section, we detail the procedure graph embedding discriminant analysis on DPS of SPD manifolds. 
As such, we propose to employ a method similar to the one discussed in \cite{shirazi2013graph}, where the difference lies on projecting the SPD matrices into DPS rather than embedding such matrices into RKHS. In the experiment section, we show that the proposed graph embedding discriminant analysis based on DPS outperforms similar methods over RKHS. 
\par
A graph \mbox{\small $\left( \Mat{V}, \Mat{E} \right)$}
in this context refers to a collection of vertices or nodes {\small $\Mat{V}$},
and a collection of edges that connect pairs of vertices.
We note that {\small $\Mat{E}$} is a symmetric matrix with elements
describing the similarity between pairs of vertices.
Moreover, the diagonal matrix {\small $\Mat{D}$} and the Laplacian matrix {\small $\Mat{L}$}
of a graph are defined as \mbox{\small $\Mat{L} = \Mat{D} - \Mat{E}$},
with the diagonal elements of \mbox{\small $\Mat{D}$}
obtained as \mbox{\small $\Mat{D}(i,i)=\sum_{j} \Mat{E}(i,j)$}.
\par
\noindent
Let $\varphi$ be the function that projects SPD matrices into DPS. As such, using (\ref{eqn:DPSMap}) :
\begin{equation}
\begin{array}{lll}
\varphi \left( \Mat{X}_i \right)\\
\mbox{=} & \Vec{k}\left(\Mat{X}_i,\mathbb{X} \right){\Mat{K}}^{\frac{1}{2}} \begin{bmatrix} \left ( \alpha \Mat{W^l} \right) & \left( \beta \Mat{W^g} \right)
\end{bmatrix}\\
\mbox{=} &
\Hat{\Vec{k}_i}
\end{array}
\end{equation}%
Given {\small $N$} labelled points {\small $\mathbb{X}= \left\{ (\Mat{X}_i, l_i) \right\}_{i=1}^{N}$}
from the underlying SPD manifold {\small $\mathcal{S}_{d}^{+}$},
where {\small $l_i \in \left\{ 1,2, \cdots, C \right\}$}; and {\small $\mathbb{\Hat{K}}= \left\{ (\Vec{\Hat{k}}_i, l_i) \right\}_{i=1}^{N}$}
from the training set on the learned DPS; the local structure of the data can be modeled by building a within-class similarity graph {\small $\Mat{E}_w$}
and a between-class similarity graph {\small $\Mat{E}_b$}. 
The simplest forms of {\small $\Mat{E}_w$} and {\small $\Mat{E}_b$}
are based on the nearest neighbour graphs defined below:
\begin{equation}
\Mat{E}_w(i,j) =
\left\{
\begin{matrix}
1, & \mbox{if} \; \Vec{\Hat{k}}_i \in N_w(\Vec{\Hat{k}}_j ) \; \mbox{~or~} \; \Vec{\Hat{k}}_j \in N_w(\Vec{\Hat{k}}_i ) \\
0, & \mbox{otherwise}
\end{matrix}
\right.
\label{eqn:Gw}
\end{equation}%
\noindent
\begin{equation}
\Mat{E}_b(i,j) =
\left\{
\begin{matrix}
1, & \mbox{if} \; \Vec{\Hat{k}}_i \in N_b(\Vec{\Hat{k}}_j ) \; \mbox{~or~} \; \Vec{\Hat{k}}_j \in N_b(\Vec{\Hat{k}}_j ) \\
0, & \mbox{otherwise}
\end{matrix}
\right.
\label{eqn:Gb}
\end{equation}%
\noindent
In \eqref{eqn:Gw},
{\small $N_w( \Vec{\Hat{k}}_i )$} is the set of $\nu_w$ neighbours
{\small $\left \{\Vec{\Hat{k}}_i^1,\Vec{\Hat{k}}_i^2,...,\Vec{\Hat{k}}_i^v \right \}$},
sharing the same label as $l_i$.
Similarly in (eqref{eqn:Gb}, {\small $N_b( \Vec{\Hat{k}}_i)$} contains {\small $\nu_b$} neighbours having different labels.

Our aim is to simultaneously maximize a measure of discriminatory power and preserve the geometry of points.
This can be formalised by finding \mbox{\small $\mathbb{W}: \varphi \left( \Mat{X}_i \right) \rightarrow \Vec{y}_i$}
such that the connected points of {\small $\Mat{E}_w$} are placed as close as possible,
while the connected points of {\small $\Mat{E}_b$} are moved as far as possible.
The mapping is sought by optimising the following two objective functions:
\begin{eqnarray}
f_1 & = & \min{\frac{1}{2}\sum\nolimits_{i,j} {\| \Vec{y}_i-\Vec{y}_j \|^2 \Mat{E}_{w}(i,j)} } \label{eqn:opt_Gw} \\
f_2 & = & \max{\frac{1}{2}\sum\nolimits_{i,j} {\| \Vec{y}_i-\Vec{y}_j \|^2 \Mat{E}_{b}(i,j)} } \label{eqn:opt_Gb}
\end{eqnarray}%
Eqn. \eqref{eqn:opt_Gw} punishes neighbours in the same class if they are mapped far away,
while Eqn. \eqref{eqn:opt_Gb} punishes points of different classes if they are mapped close together.
According to the representer theorem~\cite{Shawe-Taylor:2004:KMP}, the solution
\mbox{\small $\mathbb{W} = \left[\Vect{\gamma}_1 | \Vect{\gamma}_2 | \cdots | \Vect{\gamma}_r \right]$},
can be expressed as a linear combination of data points,
\ie, \mbox{\small $ \Vect{\gamma}_i = \sum_{j=1}^{N}{w_{i,j} \varphi \left( \Mat{{X}_j}\right) }$}:
\begin{equation}
\Vect{Y}_i = \left(
\left \langle \Vect{\gamma}_1, \varphi \left( \Mat{X}_i \right) \right \rangle ,
\left \langle \Vect{\gamma}_2, \varphi \left( \Mat{X}_i \right) \right \rangle ,
\cdots,
\left \langle \Vect{\gamma}_r, \varphi \left( \Mat{X}_i \right) \right \rangle
\right )^T
\label{eqn:RGDA_1}
\end{equation}%
Since :
\begin{equation}
\begin{array}{lll}
& \left\langle \Vect{\gamma}_l, \varphi \left( \Mat{X}_i \right) \right\rangle \\
\mbox{=} &
\sum_{j=1}^{N}{w_{l,j} (\Vec{k}\left(\Mat{X_i},\mathbb{X} \right){\Mat{K}}^{\frac{1}{2}} \begin{bmatrix} \left ( \alpha \Mat{W^l} \right) & \left( \beta \Mat{W^g} \right)
\end{bmatrix})^T} \\
\mbox{=} &
\sum_{j=1}^{N}{w_{l,j} \Vec{\hat{k_j}}}
\end{array}
\end{equation}
\noindent
,then {\small $ \Vect{Y}_i = \Mat{W}^T \Vect{\Hat{k}}_i$}, where:
\[ \Mat{W} = \begin{pmatrix}
w_{1,1} &w_{1,2} &\cdots &w_{1,r} \\
w_{2,1} &w_{2,2} &\cdots &w_{2,r} \\
\vdots &\vdots &\vdots &\vdots \\
w_{N,1} &w_{N,2} &\cdots &w_{N,r} \\
\end{pmatrix} \]
Substitution of this into \eqref{eqn:opt_Gw} results in:
\begin{equation}
\begin{array}{llll}
& \frac{1}{2} \sum_{i,j} {\| \Vect{y}_i - \Vect{y}_j \|^2 \Mat{E}_{w}(i,j)} \\
\mbox{=} & \frac{1}{2} \sum_{i,j} {\| \Mat{W}^T \Vect{\Hat{k}}_i - \Mat{W}^T \Vect{\Hat{k}}_j \|^2 \Mat{E}_{w}(i,j)} \\
\mbox{=} & \tr{ \left( \Mat{W}^T \mathbb{\Hat{K}} \Mat{D}_w \mathbb{\Hat{K}}^T \Mat{W} \right)}
- \tr{\left( \Mat{W}^T \mathbb{\Hat{K}} \Mat{E}_w \mathbb{\Hat{K}}^T \Mat{W} \right)}
\end{array}
\label{eqn:RGDA_2}
\end{equation}%
\noindent
where
\mbox{\small $\mathbb{\Hat{K}} = \left[ \Vect{\Hat{k}}_1  \Vect{\Hat{k}}_2  \cdots  \Vect{\Hat{k}}_N \right]$}.
Considering that \mbox{\small$\Mat{L}_b=\Mat{D}_b-\Mat{W}_b$}, in a similar manner
it can be shown that \eqref{eqn:opt_Gb} can be simplified to:
\begin{equation}
    \begin{array}{llll}
        & \frac{1}{2} \sum\nolimits_{i,j}{\| \Vect{y}_i-\Vect{y}_j \|^2 \Mat{E}_{b}(i,j)}\\
        = & \tr{\left( \Mat{W}^T \mathbb{\Hat{K}} \Mat{D}_b \mathbb{\Hat{K}}^T \Mat{W} \right)} -
           \tr{\left( \Mat{W}^T \mathbb{\Hat{K}} \Mat{E}_b \mathbb{\Hat{K}}^T \Mat{W} \right)}\\
        = & \tr{\left( \Mat{W}^T\mathbb{\Hat{K}}\Mat{L}_b\mathbb{\Hat{K}}^T\Mat{W} \right)}
    \end{array}
    \label{eqn:RGDA_3}
\end{equation}%

\noindent
As a result, the max versions of \eqref{eqn:opt_Gw} and \eqref{eqn:opt_Gb} can be merged by the Lagrangian method as follows:
\begin{equation}
\begin{array}{llll}
&\max \left\{\tr \left(\Mat{W}^T\mathbb{\Hat{K}}(\Mat{L}_b+\beta \Mat{E}_w)\mathbb{\Hat{K}}^T\Mat{W} \right) \right\}\\
& \text{\normalsize subject to}~~\tr \left( \Mat{W}^T\mathbb{\Hat{K}}\Mat{D}_w\mathbb{\Hat{K}}^T\Mat{W} \right) = 1
\label{eqn:RGDA_4_c}
\end{array}
\end{equation}%
\noindent
where {\small $\beta$} is a Lagrangian multiplier, and \mbox{\small$\Mat{L}_b=\Mat{D}_b-\Mat{W}_b$}.
The solution to the optimisation in (\ref{eqn:RGDA_4_c})
corresponds to the $r$ largest eigenvectors of the following generalised eigenvalue problem \cite{shirazi2013graph}:
\begin{equation}
\mathbb{\Hat{K}}\left \{\Mat{L}_b+ \beta \Mat{E}_w\right \}\mathbb{\Hat{K}}^T\Mat{W}=
\lambda \mathbb{\Hat{K}} \Mat{D}_w \mathbb{\Hat{K}}^T \Mat{W}
\label{eqn:RGDA_5}
\end{equation}%
\noindent
The experiment result indicates that the propose DPS-based Graph Embedding outperforms the RKHS-based Graph Embedding Discriminant Analysis.

\subsection{Dictionary Learning and Sparse Coding}
In this section, we describe DPS-based dictionary learning and sparse coding on SPD manifolds.  First, we project the SPD matrices into DPS through (\ref{eqn:DPSMap}), which is followed by jointly minimizing the following energy function, i.e., $min \left ( J_{\small{\mathbb{D}\mathbb{V}}} \right)$
\begin{equation}
J = \sum_{j=1}^{n} \left ( \left \| (\Vec{\Hat{k}}_j) - \sum_{i=1}^{n} \Vec{v} _{j,i} (\Mat{\Hat{D}}_i)\right \|^2 +\lambda \left \| \Vec{v}_j \right \|_1 \right )~,~\Vec{v}_j \in \mathbb{R}^n
\label{eqn:kl}
\end{equation}
, where $\Mat{X}_j \in \mathcal{S}_{d}^{+}$, and $ \Vec{\Hat{k}}_j$ is a of the SPD matrix on DPS, $\mathbb{V} = \{ \Vec{v}_1, \Vec{v}_2,~\cdots ~, \Vec{v}_n\}$ represents the sparse codes, $\Mat{\Hat{D}} = \{ \Mat{\Hat{D}}_1, \Mat{\Hat{D}}_2,~\cdots ~, \Mat{\Hat{D}}_n\} $ is a DPS-based dictionary. 
Then we employ KSVD\cite{aharon2006img} over the SPD,to solves the optimization iteratively.
Experiments indicate that the proposed DPS-based Dictionary Learning and Sparse Coding outperforms state of the art dictionary learning algorithm on SPD manifolds.

\section{Experiments and Discussion}
\label{sec:EXP}

In this section, we investigate the performance of the proposed Distance Preserved Projection on SPD manifolds  (DPS)  \footnote{We note, there are other SPD kernels that could be employed, such as Gaussian kernel \cite{arsigny2006log, jayasumana2015kernel}, or the Stein based kernel on infinite-dimensional Covariance Descriptors \cite{harandi2014bregman}. However to keep the results comparable with previous works we use Stein Divergence Kernel in this paper.} \cite{sra2011generalized,alavi2014random}, through two sets of experiments. 
\noindent
First, we evaluate the proposed DPS-based classification and compare the performance with an RKHS-based (S-SVM) \cite{harandi2014bregman}, and Random-Projection-based (ROSE)\footnote{We note that it has been shown that adding synthetic data would help to increase the classification performance in some cases ROSES\cite{alavi2014random}. However, for some applications, i.e. face recognition, it causes performance decrease \cite{alavi2014random}; as such, we choose not to add synthetic data in this work } \cite{alavi2014random} classification methods. We also examine the effect of including both local and global information in the proposed hybrid method. Thus, we refer to the hybrid-projection based DPS using (\ref{eqn:self2}), as (H-DPS). Similarly , we refer to the local-projection based DPS using (\ref{eqn:selfl}), as (L-DPS), and the global-projection based DPS as (G-DPS).
\par 
Next, we study the performance of proposed Graph Embedding Discriminate Analysis on DPS (DPS-DA) with the RKHS-based  Discriminative analysis  (RGEDA) \cite{shirazi2013graph}. This is then followed by evaluating the proposed DPS-based dictionary learning and comparing the performance with RKHS-Based dictionary learning for SPD manifolds \cite{harandi2012sparse, harandiriemannian}. 

We validate the proposed method using four datasets as follows: (1) Texture classification \cite{randen1999filtering}; (2) Face recognition \cite{phillips2000feret}; (3) Person re-identification \cite{schwartz2009learning}; and (4) Virus Texture Classification \cite{kylberg2012segmentation}.
\subsection{Classification}

\begin{figure}[!t]
  \centerline{\includegraphics[width=0.8\columnwidth,keepaspectratio]{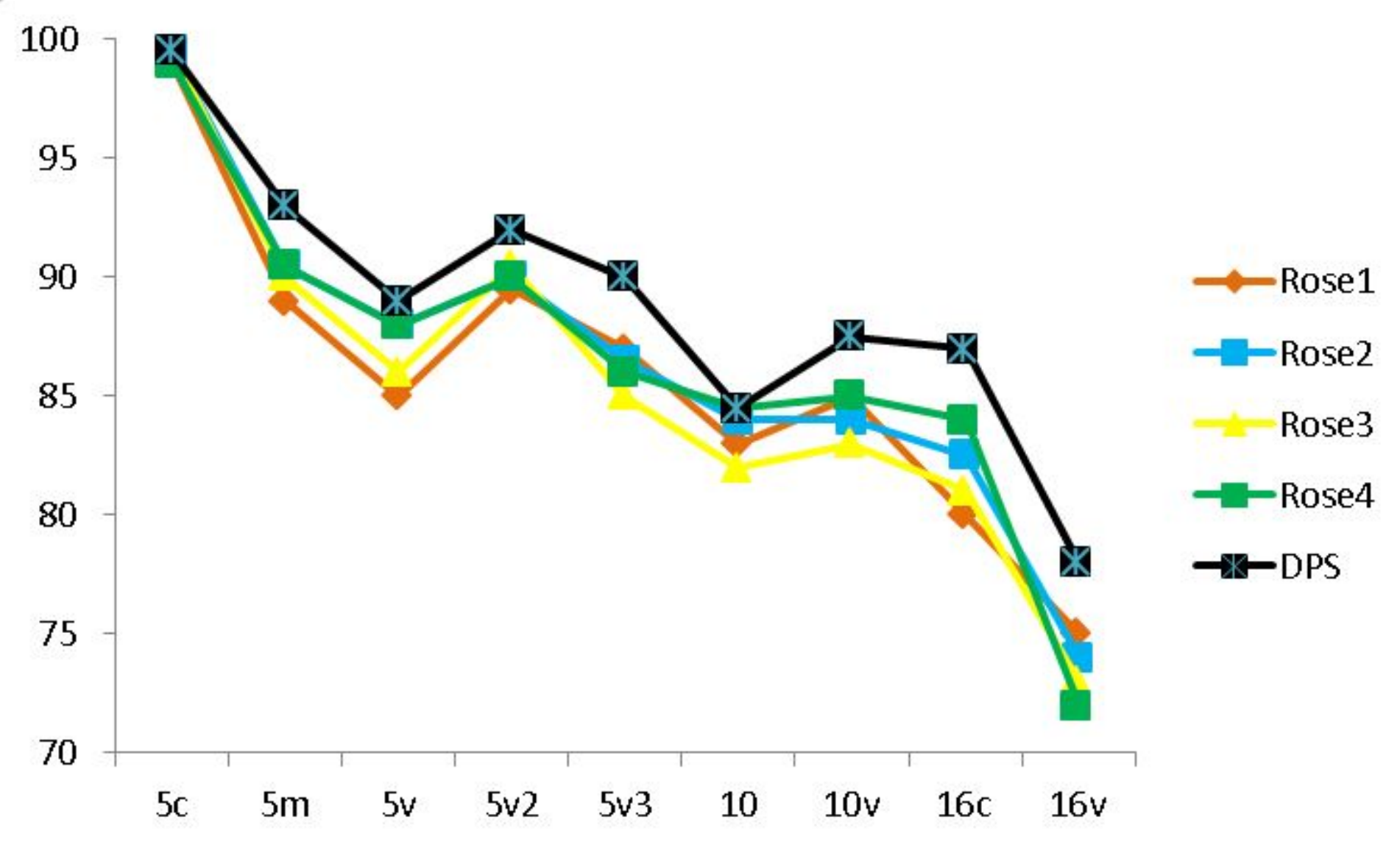}}
  \vspace{-1ex}
  \centering
  \caption
    {
    \small
    Comparing classification performance of random projection (5 runs) on the Brodatz texture dataset, with  DPS on the same dataset.
    }
  \label{fig:rl}
\end{figure}

\begin{figure}[!t]
    \begin{center}
        \begin{minipage}{1.0\columnwidth}
        \centering
            \begin{minipage}{0.30\columnwidth}
                \centerline{\includegraphics[scale = 0.15]{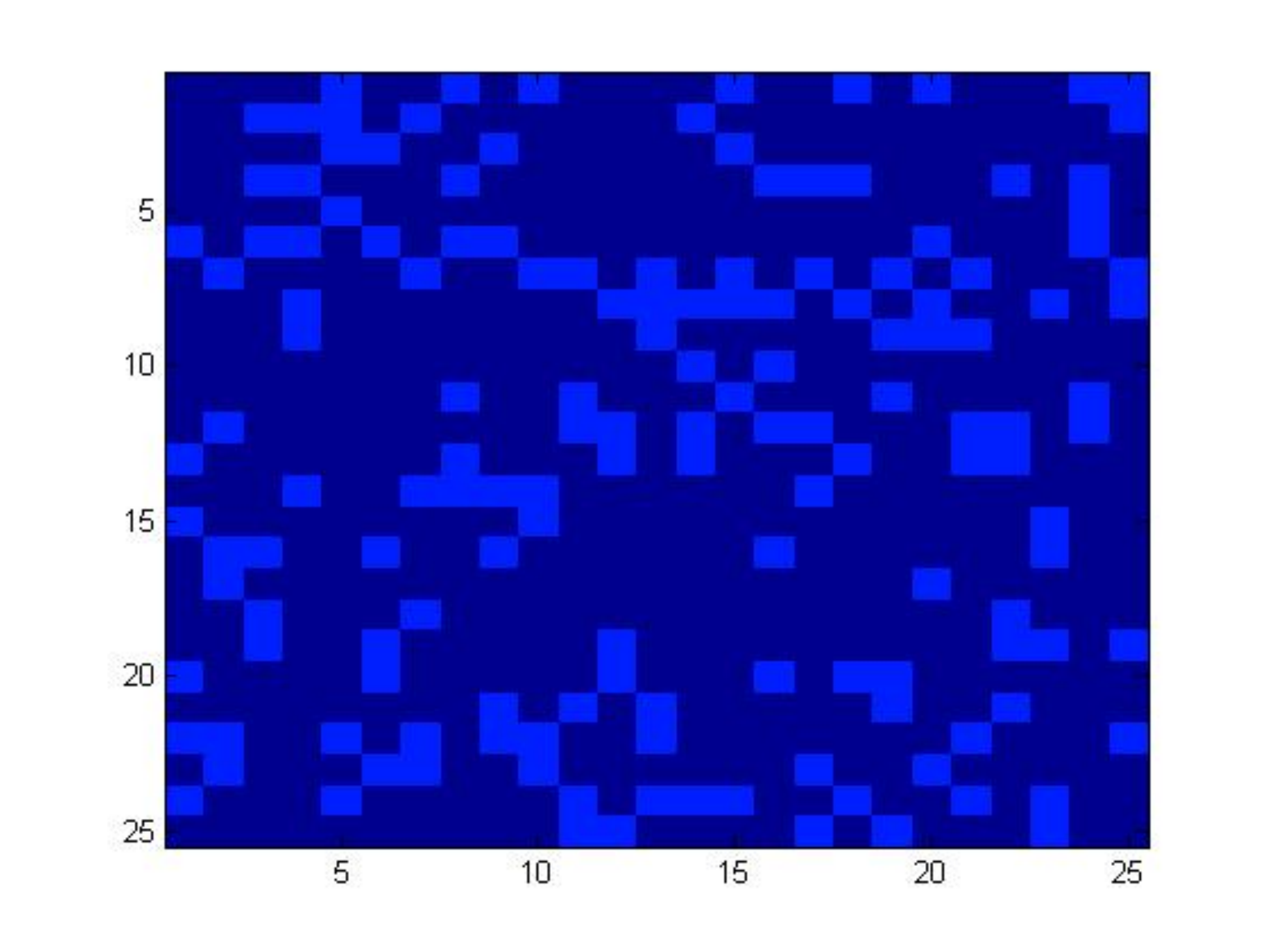}}
                \centerline{\footnotesize Rose}
             \end{minipage}
            \begin{minipage}{0.30\columnwidth}
                \centerline{\includegraphics[scale = 0.15]{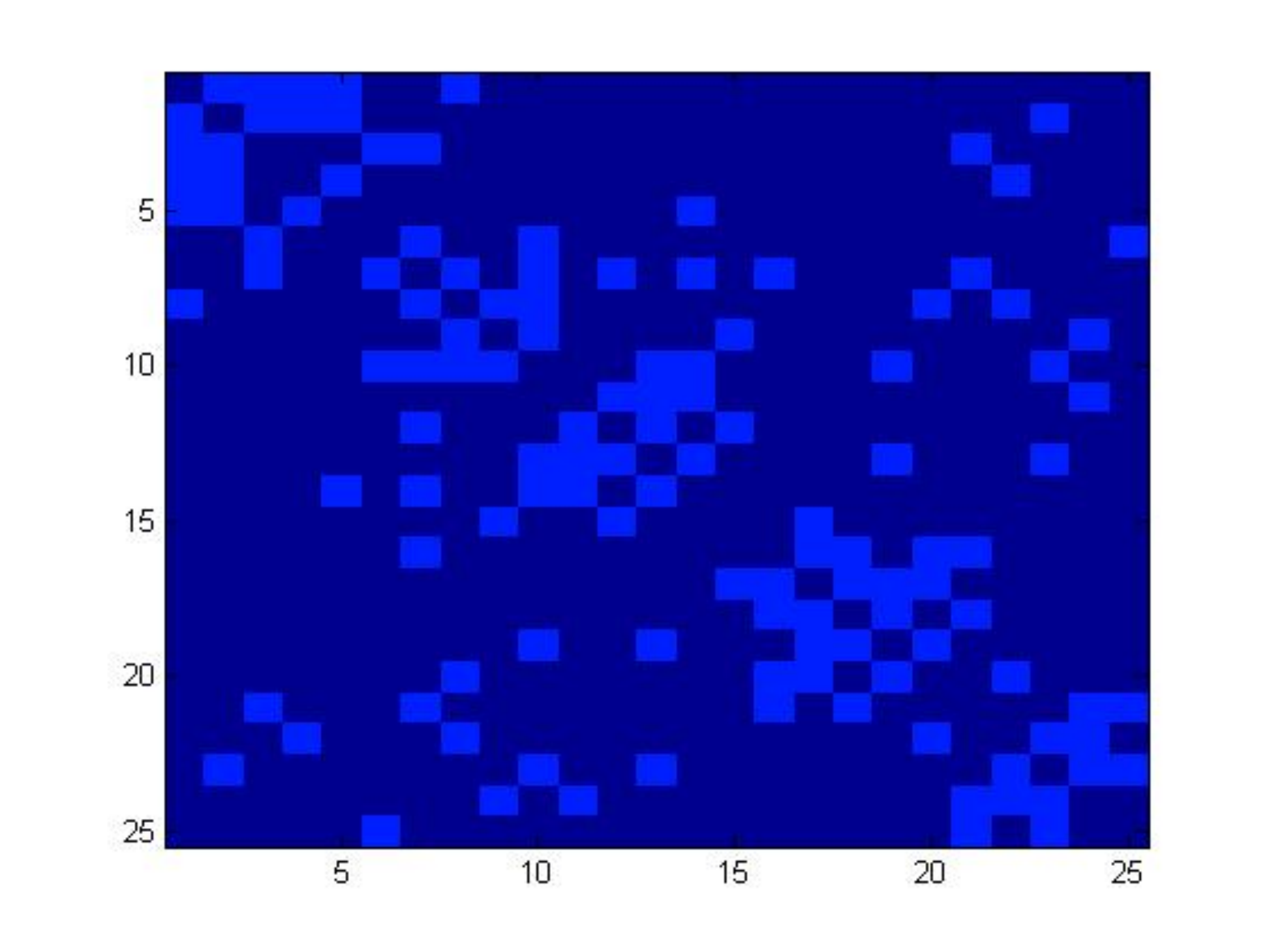}}
                \centerline{\footnotesize Local DPS}
             \end{minipage}
            \begin{minipage}{0.30\columnwidth}
                \centerline{\includegraphics[scale = 0.15]{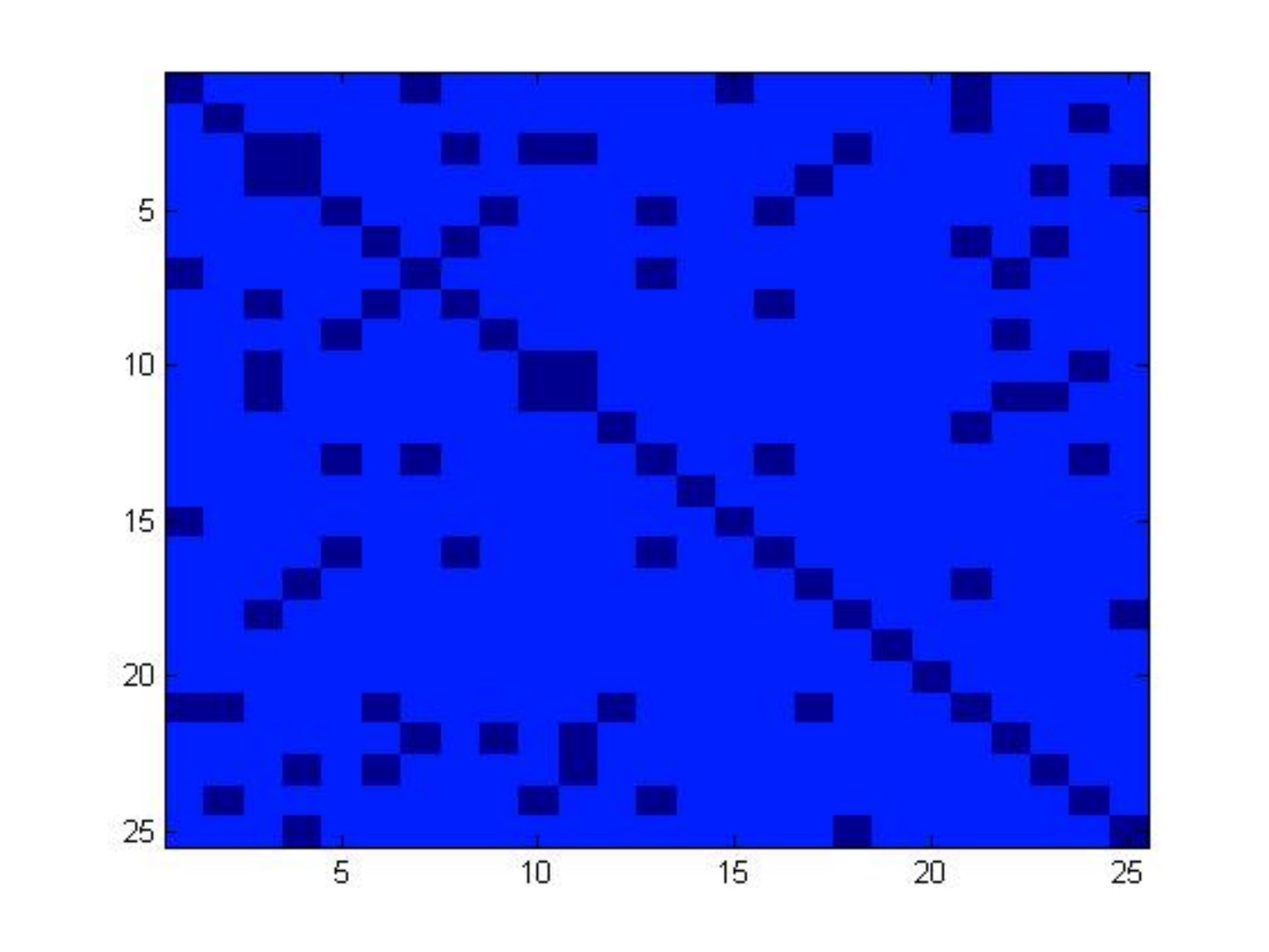}}
                \centerline{\footnotesize Global DPS}
             \end{minipage}
            
        \end{minipage}
    \end{center}
	\vspace{-1ex}
    \centering
    \caption
      {
      \small
      Examples of projection mas on sequence $5c$ of Brodatz dataset. From left to right : Projection map of Rose ($\Mat{E}$ in \ref{eqn:self}), Projection map of Local DPS ($\Mat{W^{l}}$ in \ref{eqn:Wl}), and Projection map of Global DPS ($\Mat{W^{g}}$ in  \ref{eqn:Wg}).
      }
  \label{fig:map}
\end{figure}

We start by examining the fluctuation in classification performance using ROSE. We repeat the experiment five times and show how the result varies in each run. Figure. \ref{fig:rl} illustrates the results for the texture recognition dataset (details on dataset can be found in Section. \ref{subsub:exp_texture}), and shows that the proposed SPD-based classification method outperforms ROSE in all the cases.
To provide more insight, Fig. \ref{fig:map} demonstrates examples of projection maps in ROSE and SPD.
\subsubsection{Virus Classification}
\begin{figure}[!b]
    \begin{center}
 \centerline{\includegraphics[scale = 0.45]{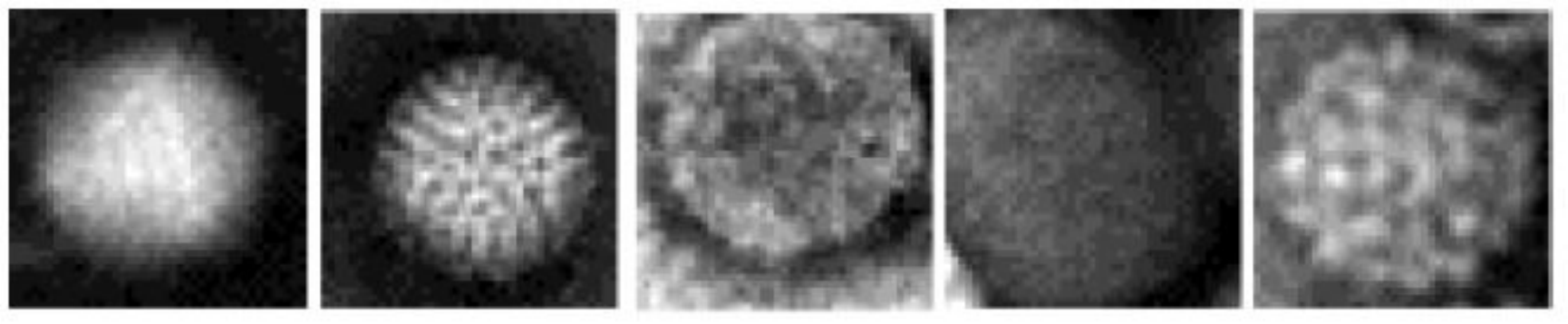}}	\vspace{-1ex}
  \end{center}
    \centering
    \caption
      {
      \small
      Examples of images from the virus dataset \cite{kylberg2012segmentation}.
      }
  \label{fig:TV_sample}
\end{figure}
The virus dataset \cite{kylberg2012segmentation} contains 15 different virus classes, where each class has hundreds of 41x41 images that were segmented automatically \cite{kylberg2012segmentation}. Samples from the virus dataset are shown in Fig. \ref{fig:TV_sample}. Following the same protocol as in \cite{harandi2012sparse, harandiriemannian} we used the 10
splits provided within the dataset in a 10 fold cross validation manner, i.e., 10 experiments with 9 splits for training and 1 split as test.
\par
At each pixel $p = (x,y)$ of an image, we then computed the 25-dimensional feature vector as following:
\begin{small}
\begin{equation}
 F_p
  \mbox{=}
  \left[
    I_p,
    \left| \frac{\partial I}  {\partial x}  \right|,  \left|\frac{\partial I}  {\partial y}  \right|,
    \left| \frac{\partial^2 I}{\partial x^2}\right|,  \left|\frac{\partial^2 I}{\partial y^2}\right|,|G_{0,0}(x,y)|,\cdots\hspace{-0.4ex},|G_{4,7}(x,y)|
  \right] \nonumber
    \label{eqn:Riemannian_KERNEL_EQN}
\end{equation}%
\end{small}
We report the mean recognition accuracies over the 10 runs. Tale. \ref{tab:TV}  shows that the proposed DPS outperforms S-SVM,  ROSE, and the PLS-based Covariance Discriminant Learning (CDL) technique \cite{wang2012covariance}, which is considered as a state of the art technique. We also observed that the generating Hybrid method advances the performance of DSP ($\%78.7$), and outperforms both global ($\%78$), and local DPS($\%77.6$).
\begin{table}[!h]
  \centering
 \small
    \begin{tabular}{lcccc}
    \toprule
 &{\bf CDL}~
    &{\bf S-SVM}~
&{\bf ROSE}~
   &{\bf H-DPS}~\\
    \toprule
     {\bf Average}       &${69.5\pm 3.1}$            &${76.5\pm 3.3}$       &${76\pm 4.1}$   &${\bf{78.7} \pm 3.2 }$\\
    \bottomrule
   \end{tabular}
  \caption
    {
    \small
    Recognition accuracy (in \%) on the Virus classification dataset}
    \label{tab:TV}
\end{table}
\subsubsection{Texture Classification}
\begin{figure}[!t]
  \centerline{\includegraphics[width=0.7\columnwidth,keepaspectratio]{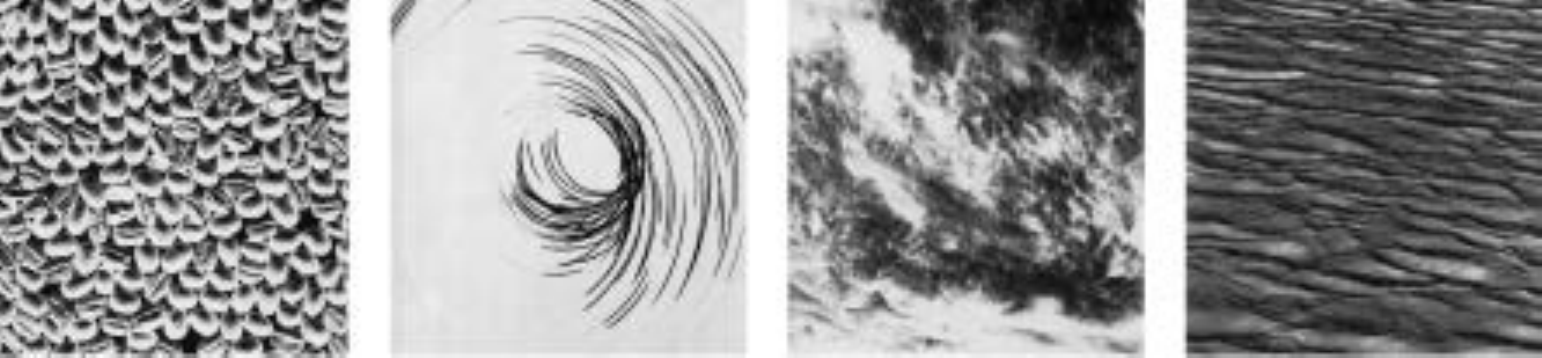}}
  \centering
  \caption
    {
    \small
    Samples of Brodatz texture dataset~\cite{randen1999filtering}.
    }
  \label{fig:brodatz}
  \end{figure}

To examine the performance of the propose DPS on classification using the Brodatz texture dataset \cite{randen1999filtering} (Examples are shown in
Fig.~\ref{fig:brodatz}, we follow the test protocol advised in~\cite{sivalingam2010tensor}.

Nine test scenarios with various
number of classes were generated.  The test scenarios included 5-texture (`5c', `5m', `5v', `5v2', `5v3'), 10-texture (`10', `10v') and 16-texture (`16c', `16v') mosaics.
To create a Riemannian manifold, we follow the same protocol as described in \cite{alavi2014random, shirazi2013graph}. As such, we first downsampel each image to $256 \times 256$, followed by splitting them into 64 regions of size $32 \times 32$.
The feature vector for each pixel $I\left(x,y\right)$ is defined as:
\begin{small}
\begin{equation}
 F(x, y)
  \mbox{=}
  \left[
    I\left(x,y\right),
    \left| \frac{\partial I}  {\partial x}  \right|,  \left|\frac{\partial I}  {\partial y}  \right|,
    \left| \frac{\partial^2 I}{\partial x^2}\right|,  \left|\frac{\partial^2 I}{\partial y^2}\right|
  \right] \nonumber
    \label{eqn:Riemannian_KERNEL_EQN}
\end{equation}%
\end{small}
\begin{table}[!h]
  \centering
 \small
    \begin{tabular}{lccc}    \toprule
    &{\bf S-SVM}~
    &{\bf ROSE}~
     &{\bf H-DPS}~\\
    \toprule
    {\bf 5c}         	&$\bf99.5$   &$\bf99.3$   	&$\bf99.5$\\
    {\bf 5m}         	&$86$  		&$90.1$   	&$\bf93$\\
    {\bf 5v}         	&$86.5$  	&$\bf91.6$ 	&$\bf91.6$\\
    {\bf 5v2}         	&$89.5$	    &$90.5$     &$\bf92$\\
    {\bf 5v3}         	&$87.5$ 	&$88.6$    	&$\bf90$\\
    {\bf 10}         	&$81.5$ 	&$\bf86.7$     &$\bf86.7$\\
    {\bf 10v}         	&$81.5$	    &$\bf88.1$   &$\bf88.1$\\
    {\bf 16c}         	&$80$ 		&$84.1$     &$\bf87$\\
    {\bf 16v}         	&$73.5$	    &$77.1$      &$\bf78$\\
    \bottomrule
   \toprule
{\bf Average}	       &$84.9$	    &$88.5$        &$\bf89.5$\\
    \bottomrule
   \end{tabular}
  \caption
    {
    \small
    Recognition accuracy (in \%) for the BRODATZ texture recognition dataset. 
task on BRODATZ datase}
    \label{tab:BRODATZ_DPS}
\end{table}
\label{subsub:exp_texture}
Tabel. \ref{tab:BRODATZ_DPS} shows that in all the sequences (DPS-based classification outperforms both S-SVM, and in all but one sequence (for which they perform on par), it outperforms ROSE. It also illustrates that on average, generating Hybrid method advances the performance of DSP, and outperforms both global ($\%88.5$) and local DPS ($\%88.1$).
\subsubsection{Face Recognition}
\begin{figure}[!b]
    \begin{center}
 \centerline{\includegraphics[scale = 0.6]{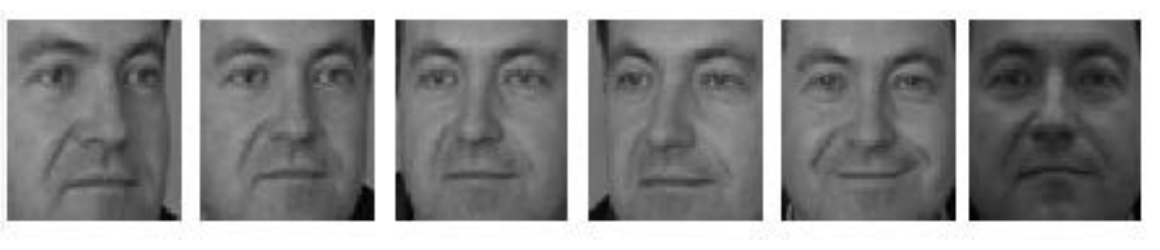}}	\vspace{-1ex}
  \end{center}
    \centering
    \vspace{-1ex}
    \caption
      {
      \small
      Examples of closely-cropped faces from the FERET 'b' subset.
      }
  \label{fig:feret_sample}
\end{figure}

For the face recognition task, we follow the same protocol as described in \cite{alavi2014random, shirazi2013graph}. We consider the subset 'b' of the FERET dataset~\cite{phillips2000feret}. This subset includes 1400 images from 198 subjects. Each image is closely cropped to merely include the face and then downsampled to $64 \times 64$. Figure \ref{fig:feret_sample} shows examples of the FERET dataset.

To evaluate the performance, we created three tests with various pose angles. In all the tests, training data consisted of the images labeled as 'bj', 'bk' and 'bf' (ie., frontal image with illumination, expression and small pose variations). Images marked as 'bd', 'be' and 'bg'(ie.,non-frontal images) were used as three separate test sets.
In our method, each face image is represented by a $43 \times 43$ covariance matrix as a point on the Riemannian manifold. To this end, for every pixel {\small $I(x,y)$}, we then computed {\small $G_{u,v}{(x,y)}$} as the response of a 2D Gabor wavelet  ($G_{u,v}(x,y)$)~\cite{lee1996image}, centered at {\small $x,y$} with orientation {\small $u$} and scale {\small $v$}.  Then the feature vector is defined as following:
\begin{equation*}
F_{x,y}
\mbox{=}
\left[~ I(x,y),~ x,~ y,~ |G_{0,0}(x,y)|,~ \cdots\hspace{-0.4ex},~ |G_{4,7}(x,y)| ~\right]
\end{equation*}%
Tabel. \ref{tab:FERET_ROSE} illustrates the performance of DPS on FERET face recognition dataset, and shows that DPS outperforms both S-SVM and ROSE. In addition, it shows that in all the cases hybrid ($\%91.5$) and global methods perform on par, and outperform the local ($\%91$) DPS. 
\label{exp_face_recognition}
\begin{table}[!h]
  \centering
 \small
    \begin{tabular}{lccccc}
    \toprule
    &{\bf S-SVM}~
    &{\bf ROSE}~
     &{\bf H-DPS}~\\
    \toprule
    {\bf bg}    &${69.5}$     &${80}$     &${\bf 85}$\\
   {\bf bf}    &$  95$         &$  95$   &${\bf 98}$\\
   {\bf be}    &$ {82}$       &$ {85}$   &${\bf94}$\\
    {\bf bd}     &${52}$        &${70.5}$     &${ \bf 75.5}$\\
    \bottomrule
       \toprule
	{\bf Average}	       &$74.6$	    &$82.6$      &$\bf91.5$\\
    \bottomrule
   \end{tabular}
  \vspace{0.5ex}
  \caption
    {
    \small
    Recognition accuracy (in \%) for the face recognition task on the ‘b’ subset of the
FERET dataset}
    \label{tab:FERET_ROSE}
\end{table}

\subsubsection{Person Re-identification}
\label{exp_reidentification}

\begin{figure}[!t]
  \centerline{\includegraphics[width=0.7\columnwidth,keepaspectratio]{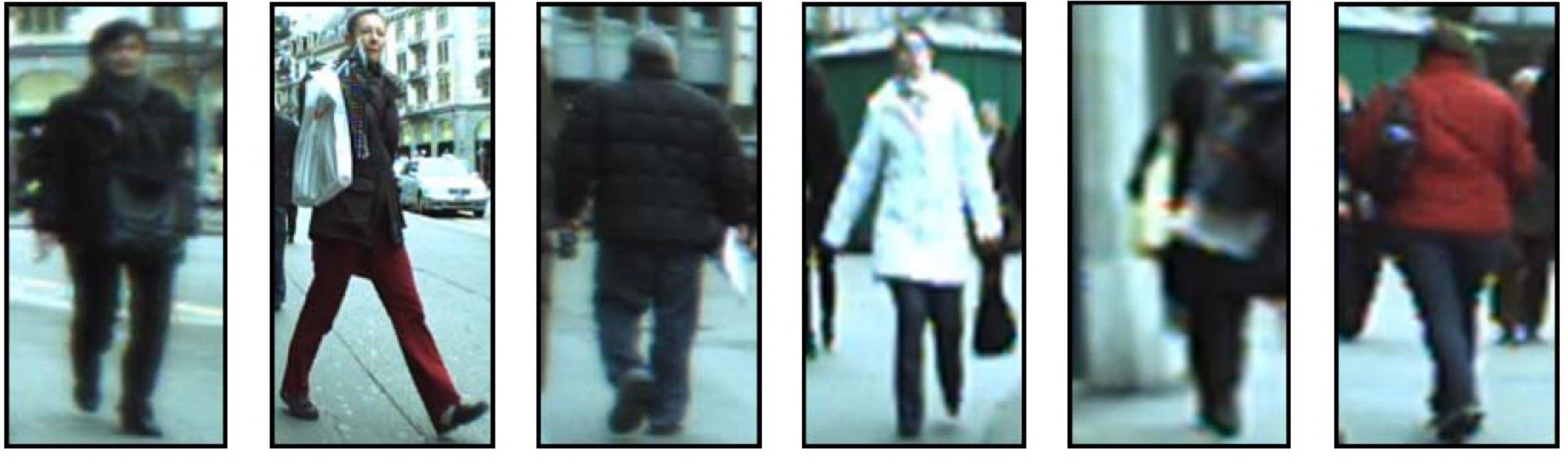}}
  \vspace{-1ex}
  \centering
  \caption
    {
    \small
    Examples of pedestrians in the ETHZ dataset.
    }
  \label{fig:ethz}
\end{figure}

In this section we test the performance of the proposed DPS method for person
reidentification task on the modified ETHZ dataset~\cite{schwartz2009learning}, and follow the same protocol described in \cite{alavi2014random, shirazi2013graph}. The original version of this dataset was captured from a moving camera~\cite{ess2007depth}, and it has been used for human detection. The main challenging aspects of ETHZ dataset are variations in pedestrians appearances and occlusions. Some sample images of the ETHZ dataset are shown in figure~\ref{fig:ethz}.
This dataset contains three video sequences. We downsampled all the images to $64 \times 32$. For each subject, the training set consists of 10 randomly selected images and the rest used for the test set. To generalise the practical assessment of the algorithm, random selection of the training and testing data was repeated 20 times.

\begin{table}[!b]
  \centering
 \small
    \begin{tabular}{lccc}
    \toprule
    &{\bf S-SVM}~
    &{\bf ROSE}~
     &{\bf H-DPS}~\\
    \toprule
    {\bf Seq.1}          &$84.4 \pm 1$  &${90.7 \pm 0.9}$   &${\bf 90.7 \pm 0.9}$\\
    {\bf Seq.2}          &$84.2 \pm 1.3$  &${91.5 \pm 1.7}$   &${\bf 92.7 \pm 1.5}$\\
    \bottomrule
       \toprule
	{\bf Average}	       &$84.3$	    &$91.2$    &$\bf91.7$\\
    \bottomrule
   \end{tabular}
  \vspace{0.5ex}
 \caption
   {
    \small
    Recognition accuracy (in \%) for the person re-identification task on Seq.1 and Seq.2 of the ETHZ dataset
    }
    \label{tab:ETHZ_DPS}
\end{table}

To create points on the Riemannian manifold, a feature vector was formed for each pixel using the position of the pixel ($x$ and $y$), the corresponding colour information ($R_{x,y}$, $G_{x,y}$ and $B_{x,y}$) and the gradient and Laplacian for colour C, defined as {\footnotesize $C_{x,y}' \mbox{=} \left[ \left|{\partial C} \middle/ {\partial x}\right|, \left|{\partial C} \middle/ {\partial y}\right| \right]$}
and
{\footnotesize $C_{x,y}'' \mbox{=} \left[ \left|{\partial^2 C} \middle/ {\partial x^2}\right|, \left|{\partial^2 C} \middle/ {\partial y^2}\right| \right]$},
respectively. Then the representation for each image is the covariance matrix using the following feature:
\begin{equation*}
F_{x,y}
\mbox{=}
\left[~
  x,~ y,~
  R_{x,y},~   G_{x,y},~   B_{x,y},~
  R_{x,y}',~  G_{x,y}',~  B_{x,y}',~
  R_{x,y}'',~ G_{x,y}'',~ B_{x,y}''~
\right]
\end{equation*}%
Table \ref{tab:ETHZ_DPS} shows that on average, DPS outperforms both  S-SVM and ROSE. We also observed that the hybrid DPS ($\%91.7$)advances the method and outperforms both local ($\%91.5$)and global DPS ($\%91$).

\subsubsection{Further Analysis}

To further examine the performance of the proposed DPS, we compare the proposed DPS-based graph embedding discriminant analysis (DPS-DA) with the similar RKHS-based method (RGEDA)\cite{shirazi2013graph}. In addition we compare the proposed DPS-based dictionary learning and sparse coding (DPS-DI) method with kernel-based dictionary learning and sparse coding on SPD manifold (RSR) \cite{ harandi2012sparse}.  \cite{harandiriemannian}. 
\begin{table}[!h]
  \centering
 \small
    \begin{tabular}{lccccc}
    \toprule
    &{\bf TSC}~
    &{\bf RGEDA}~
   &{\bf DPS-DA}~
   &{\bf RSR}~
   &{\bf DPS-DI}~\\
    \toprule
    {\bf 5c}         	&$\bf100$  &$99$        &$\bf100$  &$\bf99.5$  &$\bf99.5$\\
    {\bf 5m}         	&$75$  	    &$83$       &$\bf91.5$ &$84$  	   &$\bf91.5$\\
    {\bf 5v}         	&$82.5$  	&$83$       &$\bf91.5$   &$\bf89$    &$\bf89$\\
    {\bf 5v2}         	&$82$	    &$85$       &$\bf92$   &$89.5$	   &$\bf91.5$\\
    {\bf 5v3}         	&$81.5$ 	&$82$       &$\bf90$   &$86$       &$\bf89$\\
    {\bf 10}         	&$81$    	&$83$       &$\bf85$   &$\bf83$    &$\bf83$\\
    {\bf 10v}         	&$68$	    &$81$       &$\bf89$   &$86.5$	   &$\bf87$\\
    {\bf 16c}         	&$77$    	&$82$       &$\bf87.5$ &$82$       &$\bf84$\\
    {\bf 16v}         	&$66$	    &$77.5$     &$\bf78.5$ &$77$	   &$\bf79$\\
    \bottomrule
   \toprule
{\bf Average}	       &$79.4$	    &$84$      &$\bf89.5$     &$86.3$     &$\bf88.2$ \\
    \bottomrule
   \end{tabular}
  \caption
    {
    \small
    Recognition accuracy (in \%) for the BRODATZ texture recognition dataset.}
    \label{tab:BRODATZ_further}
\end{table}
Table. \ref{tab:BRODATZ_further} shows that on average DPS-based discriminative analysis out performs RKHS-based discriminative analysis in all the sequences of the BRODATZ dataset. It performs on par with TSC, which is considered as state of the art method, on $5c$ sequence, and outperforms TSC on all the other sequences. On average DPS-based dictionary learning and sparse coding outperforms both TSC and RSR.
\begin{table}[!h]
  \centering
 \small
    \begin{tabular}{lccccc}
    \toprule
    &{\bf SDALF}~
    &{\bf RGEDA}~
   &{\bf DPS-DA}~
   &{\bf RSR}~
   &{\bf DPS-DI}~\\
    \toprule
    {\bf Seq.1}          &$83.4$  &${90.5}$   &${\bf92}$  &${91}$ &${\bf 92}$\\
    {\bf Seq.2}          &$83.4$  &${92}$     &${\bf93.5}$    &${92}$ &${\bf 92.5}$\\
    \bottomrule
       \toprule
		{\bf Average}	       &$83.4$	    &$91.3$      &${\bf92.7}$    &$91.5$   &${\bf 92.2}$\\
    \bottomrule
   \end{tabular}
  \vspace{0.5ex}
 \caption
   {
    \small
    Recognition accuracy (in \%) for the person re-identification task on Seq.1 and Seq.2 of the ETHZ dataset
    }
    \label{tab:ETHZ_further}
\end{table}
Table. \ref{tab:ETHZ_further} shows that on average DPS-based discriminative analysis out performs RGDA, RKHS-based discriminative analysis, on ETHZ dataset. IT also outperforms SDALF in all the cases DPS outperforms SDALF \cite{SDALF_CVPR2010}, which is considered to be the state of the art. It also illustrates that DPS-based dictionary learning and sparse coding outperforms RSR and SDALF in all the cases.
\begin{table}[!h]
  \centering
 \small
    \begin{tabular}{lccccc}
    \toprule
 &{\bf CDL}~
    &{\bf RGEDA}~
   &{\bf DPS-DA}~
   &{\bf RSR}~
   &{\bf DPS-DI}~\\
    \toprule
       {\bf Average}    &${69.5}$            &${77.9}$       &${\bf{78.7}}$       &${78 }$   &${\bf{79}}$\\
    \bottomrule
   \end{tabular}
  \caption
    {
    \small
    Recognition accuracy (in \%) for the Virus Texture Recognition dataset}
    \label{tab:TV_Further}
\end{table}
Table. \ref{tab:TV_Further} shows that on average the DPS-based discriminative analysis outperforms RGEDA, CDL \cite{wang2012covariance} which is considered as state of the art , on the Virus Texture dataset. It also illustrates that DPS-based dictionary learning and sparse coding outperforms both RSR and CDL.

\section{Conclusion}
\label{sec:Conc}
In this paper, we propose to learn the optimum kernel-based projection that maps SPD matrices into a Distance Preserving Projection Space DPS), through constructing both local and global sparse similarity graphs, whose nodes are connected to each other iff they correspond to points from the same subspace. 
We show that a such mapping mitigates the drawbacks of the random projection on SPD manifolds method. Experiments on several vision tasks (person re-identification, face recognition, texture recognition, and virus classification), show that the proposed approach outperforms ROSE and several other state-of-the-art methods. We further adopt discriminative analysis (DA), and dictionary learning (DL) on the learned DPS of such manifolds, and show that such methods outperform state of the art methods on SPD manifolds.


{\small
\bibliographystyle{ieee}
\bibliography{egbib}
}

\end{document}